\documentclass{article} 
\usepackage{iclr2021_conference,times}

\usepackage{amsmath,amsfonts,bm}

\def\eqref#1{equation~\ref{#1}}

\def\1{\bm{1}}

\DeclareMathAlphabet{\mathsfit}{\encodingdefault}{\sfdefault}{m}{sl}
\SetMathAlphabet{\mathsfit}{bold}{\encodingdefault}{\sfdefault}{bx}{n}

\usepackage{hyperref}
\usepackage{url}
\usepackage[final]{graphicx}

\title{Multilingual Medical Question Answering and Information Retrieval for Rural Health Intelligence Access}

\author{Vishal Vinod*, Susmit Agrawal*, Vipul Gaurav\thanks{Equal contribution} , Pallavi R. \& Savita Choudhary \\
Sir M. Visvesvaraya Institute of Technology\\
Bengaluru, India\\
\texttt{pro.vishalvinod@gmail.com} \\
\texttt{\{susmit600, vipul\_1mv16cs124, pallavi\_cs, savitha\_cs\}@sirmvit.edu} \\
}

\iclrfinalcopy 
\begin{document}

\maketitle

\begin{abstract}
In rural regions of several developing countries, access to quality healthcare, medical infrastructure, and professional diagnosis is largely unavailable. Many of these regions are gradually gaining access to internet infrastructure, although not with a strong enough connection to allow for sustained communication with a medical practitioner. Several deaths resulting from this lack of medical access, absence of patient's previous health records, and the unavailability of information in indigenous languages can be easily prevented. In this paper, we describe an approach leveraging the phenomenal progress in Machine Learning and NLP (Natural Language Processing) techniques to design a model that is low-resource, multilingual, and a preliminary first-point-of-contact medical assistant. Our contribution includes defining the NLP pipeline required for named-entity-recognition, language-agnostic sentence embedding, natural language translation, information retrieval, question answering, and generative pre-training for final query processing. We obtain promising results for this pipeline and preliminary results for EHR (Electronic Health Record) analysis with text summarization for medical practitioners to peruse for their diagnosis. Through this NLP pipeline, we aim to provide preliminary medical information to the user and do not claim to supplant diagnosis from qualified medical practitioners. Using the input from subject matter experts, we have compiled a large corpus to pre-train and fine-tune our BioBERT based NLP model for the specific tasks. We expect recent advances in NLP architectures, several of which are efficient and privacy-preserving models, to further the impact of our solution and improve on individual task performance.
\end{abstract}

\section{Introduction}

Several rural regions have a severe shortage of doctors \citep{sharma2015india}. Even in places where a practitioner is available, they have to attend to an abnormally large number of patients daily, constraining the quality of the diagnosis and resulting in severely overworked practitioners. In several regions, patients need to travel long distances to neighboring villages even for simple and common consultations. For long-standing health conditions, medical practitioners require a detailed health record for the patient, which is more often than not unavailable or completely absent. Many of the health conditions, deficiencies, and symptoms can be identified early, and preventable treatment can be undertaken at early stages to avert several deaths stemming from the lack of awareness. The growing internet penetration rates are a big positive since it allows patients from even remote regions to access information on the internet, although still constrained by the language barrier in many regions. Thus, there is a need for a solution that is easily accessible, that provides accurate medical information in indigenous languages, can store information from previous doctor interactions and provide advisory information and as a result reduce the accessibility problem in rural healthcare.

Access is a challenging problems to solve primarily because of the various privacy and ethical concerns surrounding a medical application that provides information as a question-answer system and stores sensitive medical information from previous interactions with a doctor (to aid their diagnosis of the individual). We are transparent with our aim that a first-point-of-contact medical advisory assistant cannot, and will not supplant the diagnosis and advice from a trained medical practitioner. We only aim to reduce the information accessibility gap by leveraging the latest advances in NLP and Deep Learning to design a medical assistant that can be used offline, is resource efficient, multilingual (for vernacular languages), and is accurate in its associated tasks such as NER (Named Entity Recognition), RE (Relation Extraction), QA (Question Answering), NMT (Neural Machine Translation) and generative pre-training for query processing and providing a human-like final response. The medical corpus and the BERT-based model for question-answering and encoding onto an embedding space must be effectively trained with a large and reliable corpus of trusted medical information. We believe that such a dialogue representation model can be an excellent enabler for rural regions to improve their understanding of diseases. Accurate access to health information in vernacular languages can prevent avoidable deaths by identifying problems early.

\section{Related Work}

Advances in NLP have been progressing rapidly towards model architectures that are domain and task agnostic leading to a general language model that can be pre-trained and fine-tuned for downstream domain-specific tasks. This progress has been influenced by the seminal works on sequence-to-sequence models \citep{sutskever2014sequence}, transformer models \citep{vaswani2017attention}, and memory architectures. More recently, the BERT (Bi-directional Encoder Representation for Transformer) \citep{devlin2018bert} masked language model (MLM) has become one of the popular NLP architectures for NER (Named Entity Recognition), relation extraction, and question-answering. BERT is a general MLM that can be further pre-trained for domain-specific language representation and fine-tuned for specific tasks. The BioBERT architecture \citep{lee2020biobert} is a contextualized MLM that is a pre-trained BERT on a biomedical corpus. The pre-training leverages the general language representations in BERT that is transferable across domains. BioBERT is pre-rained on PubMed abstracts (4.5 Billion words) and PMC articles (13.5 Billion words), which makes it a strong candidate for our question-answer embedding and information retrieval use-case.

The medical reports and clinical data are much more complicated to general medical terms and features included in the deep representations of the clinical medical text and embedding. ClinicalBERT \citep{huang2019clinicalbert} identifies the unique sequence of tokens and has a self-attention mechanism that outperforms BioBERT in modeling notes from medical practitioners. Such an implementation would be useful for our application to store information regarding previous interactions with medical practitioners. XLNET \citep{yang2019xlnet} showed that the capability of bi-directional modeling of contexts and denoising auto-encoding based pretraining like BERT achieves better performance over pretraining approaches based on auto-regressive language modeling. XLNET enables learning bi-directional contexts by maximizing the expected likelihood of all permutations of the factorization order. Although XLNET outperforms BERT-based models on several language tasks, we have used the BioBERT architecture owing to its ease-of-use and the option to further fine-tune on tasks required for our architecture. We use the GPT-2 architecture \citep{radford2019language} for generative pretraining to generate natural language answers from our information retrieval component.
	 
LABSE (language-agnostic BERT Sentence Embed-ding) \citep{feng2020language} is a BERT architecture trained as an MLM (Masked Language Model) and a TLM (Translation Language Model) on a translation ranking task to perform language translation with a 500k token vocabulary onto 109 different languages. The input and target are encoded using a shared transformer embedding, forcing both to equal representation. Prior work \citep{olvera2011multilingual, lai2018review, daniel2019towards, arivazhagan2019massively} has discussed the applicability of English for query processing and hence we use LABSE only for input and output query translation. This multilingual architecture can enable inclusive healthcare for all and improve patient participation in healthcare. Several other more recent architectures such as GPT-3 \citep{brown2020language}, Reformer \citep{kitaev2020reformer}, DistillBERT \citep{sanh2019distilbert}, and PRADO \citep{kaliamoorthi2019prado} show great promise in terms of efficient language modeling and performance on QA and information retrieval when made fully accessible. The DistillBERT model can be used for extractive summarization. 

\section{Methodology}

\begin{figure}[h]
\begin{center}
\fbox{\rule[-.5cm]{0cm}{4cm} \includegraphics[scale=0.58]{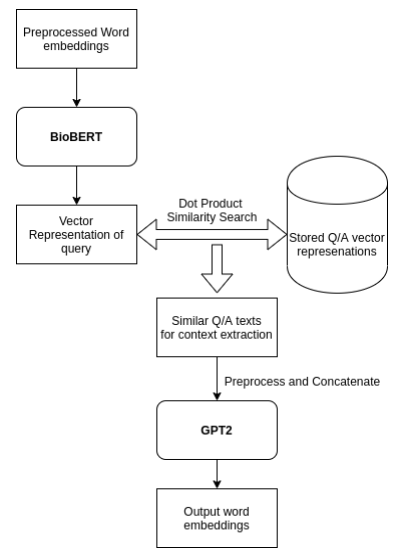}}
\fbox{\rule[-.5cm]{0cm}{4cm} \includegraphics[scale=0.35]{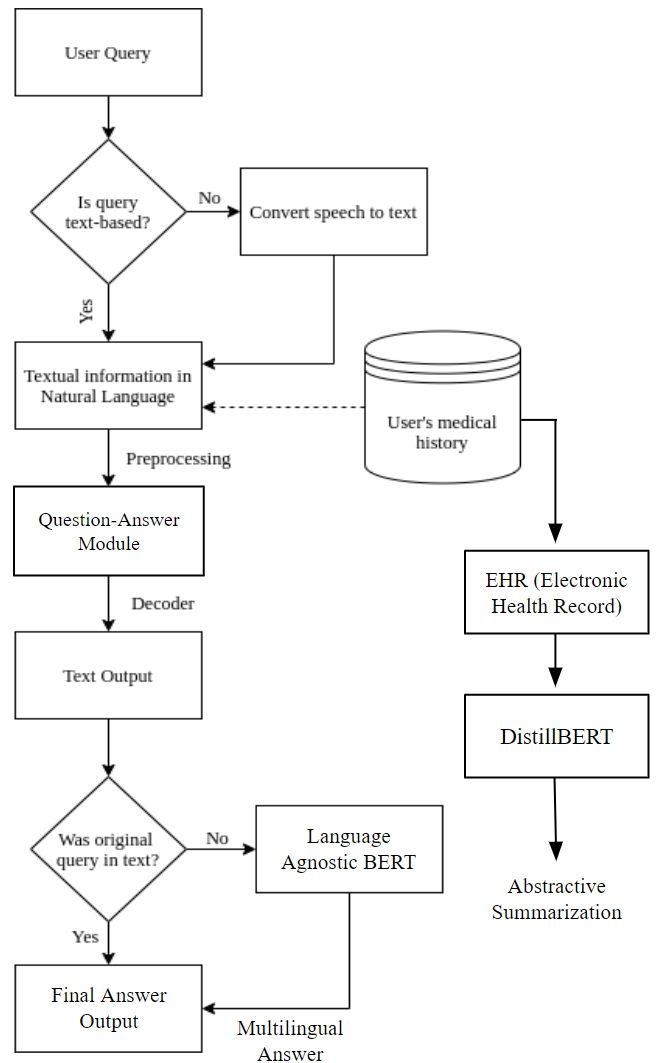}}
\end{center}
\caption{(a) Proposed architecture of the medical QA system; (b) The Question-Answer module.}
\end{figure}

The proposed model architecture builds on the BioBERT \citep{lee2020biobert}, LABSE \citep{feng2020language}, and GPT-2 \citep{radford2019language} model architectures for multi-lingual question answering and information retrieval. We use the LABSE model over NMT (Neural Machine Translation) because LABSE is a BERT based architecture an can be pre-trained on biomedical corpora to effectively translate medical queries keeping the domain-specific context for several tasks. LABSE is available for easy use from TensorFlow Hub. The model architecture has five main stages:

\textbf{Compile the dataset}. The large dataset of question-answer pairs are for reference to the model to be able to generalize well even in cases of lack of information. These pairs provide the model with additional context, which becomes necessary when patients cannot fully and accurately describe their condition and symptoms. The proposed architecture also has an EHR (Electronic Health Record) summarizer that takes the patient’s previous diagnosis and status information as input and summarizes any prevailing medication or treatment. This module uses the DistillBERT abstractive summarizer fine-tuned on Springer Nature Digital Medicine abstracts.

The Question-Answer pairs dataset is compiled from forums such as Reddit, WebMD, and Question Doctors and are compiled to create a dataset consisting of 325,000 unique pairs. The question-answer pairs are encoded onto an intermediate embedding space using the BioBERT model onto a Dense layer encoding, which can then be used for information retrieval based on a similarity-search.

\textbf{Text processing pipeline}. Creating a text pre-processing pipeline using a corpus of words; the system needs a predetermined set of words that it can use to contextualize the meaning of a query in natural language. Several such public datasets are available. However, our work requires that the corpus used must contain a large set of clinical and biomedical terms in its vocabulary. For this purpose, we use the PubMed dataset. PubMed is primarily based on the MEDLINE database of references and abstracts on life sciences and biomedical topics. The stage involves the following steps: text tokenization, creating embeddings for the tokens, and creating special tokens such as EOS and UNK, as well as their embeddings. Word embeddings are then created out of the words in vocabulary. Each em-bedding is assigned a positive integer ID. Once the sentence is obtained, we use the LABSE model to translate the response to the English language for the BioBERT model.

\textbf{Training the model architecture}.  We initialize the BERT model with BioBERT weights. During training, we use the question and answer pairs from the compiled train dataset above to obtain their intermediate representations in embedding space. We share the weights for both question head as well as the answer head borrowing the model architecture from DocProduct \citep{gupta2019docproduct}. These embeddings are then encoded onto separate Dense layers, one each for the question and answer head, used later for similarity lookup. 

\textbf{Generating natural-language answers}. We use the GPT-2 (117M) parameter architecture to generate natural language answers to our retrieved answer by bolstering it with relevant text. During training, the retrieved queries are pre-processed and input to the GPT-2 model to compute the loss to train the embedding heads. Same as DocProduct, we use the dot product of the question-answer embedding pair from the similarity lookup and then compute the Softmax of the rows for each question. The loss is comput-ed using cross-entropy, and then gradients are propagated to train the model layers. We expect the dense embedding layers to capture the similarity of the question-answer pairs.

\textbf{Conveying the results}. In this stage, we use the output from GPT-2, which is the bolstered similar question-answer text obtained from our inference time similarity lookup. This is again processed by the LABSE model to obtain the answer in the required language to provide to the user. Additionally, we also use the extractive summaries of previous doctor consultations and use the DistillBERT model to embed the sentences, cluster them, and then to find sentences that are closest to the cluster centers. These mini-summaries can be useful for doctors to provide more in-depth diagnostics to the patient. (Note: This part of the architecture is still under active development)

\begin{table}[t]
\label{table:results}
\begin{center}
\begin{tabular}{ccccc}
\multicolumn{1}{c}{\bf Dataset} &\multicolumn{1}{c}{\bf BERT} &\multicolumn{1}{c}{\bf BioBERT} &\multicolumn{1}{c}{\bf BioBERT}&\multicolumn{1}{c}{\bf Proposed} \\
 &(WIKI+Books) &(+PubMed) &(+PMC) &(+PubMed) \\
\hline \\
BioASQ 4b  & \underline{27.33} &25.47 &26.09 &\textbf{27.78} \\ [0.2ex]
BioASQ 5b  &39.33 &41.33 & \underline{42.00} &\textbf{42.61} \\ [0.2ex]
BioASQ 6b  &33.54 &\textbf{43.48} &41.61 & \underline{42.38} \\ [1ex]
\hline
\end{tabular}
\caption{Strict Accuracy (S) reported for the respective datasets. The best scores are in bold and the second best scores are underlined.}
\end{center}
\end{table}

\section{Results and Discussion}

The outcome of this work is an application of dialogue representation system designed to provide quality medical consultations solving the problem of first point contact of patients with doctors. Our work obtains promising results for biomedical question answering (in English) for the BioASQ 4b. BioASQ 5b and BioASQ 6b datasets with Strict Accuracy (S) of 27.78, 42.61 and 42.38 respectively with only the PubMed dataset and BioBERT model as tabulated in Table 1. Our work can be implemented as an API endpoint for quick access to question-answers and user specific account information. We have also worked on making the model efficient with model pruning, post-training quantization and NNAPI delegates for deployment on Android smartphones with tflite acceleration for optimized offline performance. The promising QA results in English and out-of-the-box translation wrappers for input speech-to-text and output text translation powered by LABSE enables wide adoption of the multilingual dialogue model especially in rural health intelligence. 

With the COVID-19 induced digital adoption across the world, rural populations are most affected by the information access gap. We expect our solution to be a small contribution toward empowering people through advances in Machine Intelligence and NLP to get up-to-date-information. NLP is evolving to help the market of healthcare substantially to improve personalized solutions such as the work by \cite{bornea2020multilingual}, and with the recent advancements in training language models with LABSE, BioBERT, and GPT-2, the ability of QA to understand the human queries.

\bibliography{iclr2021_conference}
\bibliographystyle{iclr2021_conference}


\end{document}